%% file: main.tex
\documentclass[]{alaya}
\usepackage{makecell}
\usepackage{wrapfig}
\usepackage{tabularx}
\usepackage{textcomp}
\usepackage{stfloats}
\usepackage{url}
\usepackage{verbatim}
\usepackage{titlesec}
\usepackage{adjustbox}
\usepackage{multirow}
\usepackage{pifont}
\usepackage[sc]{mathpazo}
\usepackage{tikz}
\usepackage{amsmath,amssymb}
\usepackage{colortbl}
\usepackage[numbers,sort&compress]{natbib}
\usepackage{booktabs}
\usepackage{hyperref}
\usepackage{graphicx}
\RequirePackage{xspace}
\makeatletter
\DeclareRobustCommand\onedot{\futurelet\@let@token\@onedot}
\def\@onedot{\ifx\@let@token.\else.\null\fi\xspace}
\usepackage[most]{tcolorbox}
\usepackage{array}
\usepackage{siunitx}
\usepackage{caption}
\definecolor{headerpurple}{HTML}{d8d2fc}
\definecolor{rowgray}{gray}{0.95}
\usepackage{CJKutf8}

\makeatother

\definecolor{adptorange}{RGB}{248, 205, 172}
\definecolor{cmpblue}{RGB}{189, 215, 238}

\definecolor{our_red}{RGB}{232,157,160}
\definecolor{our_blue}{RGB}{136,206,230}
\definecolor{our_orange}{RGB}{246,200,168}
\definecolor{our_green}{RGB}{178,211,164}

\definecolor{mygray}{HTML}{f0f0f0}

\usepackage{bbding}
\usepackage{fontawesome}
\usepackage{float}
\usepackage{flafter}
\usepackage{needspace}
\usepackage{algorithm}
\usepackage{algorithmic}
\usepackage{booktabs}
\usepackage[table]{xcolor}
\usepackage{listings}
\definecolor{lstbg}{gray}{0.96}
\definecolor{lstcomment}{rgb}{0.35,0.45,0.40}
\newlength\savewidth

\newcolumntype{x}[1]{>{\centering\arraybackslash}p{#1pt}}
\newcolumntype{y}[1]{>{\raggedright\arraybackslash}p{#1pt}}
\newcolumntype{z}[1]{>{\raggedleft\arraybackslash}p{#1pt}}

\renewcommand{\paragraph}[1]{\vspace{1.25mm}\noindent\textbf{#1}}

\definecolor{green}{HTML}{009000}
\definecolor{red}{HTML}{ea4335}


\title{StreamDuo: A Live Commentator and a Research Agent for Unseen Games}

\author[1,2]{\mbox{Zixian Gao}}
\author[3]{\mbox{Penghao Luo}}
\author[1]{\mbox{Xiangcheng Cao}}
\author[2]{\mbox{Lixin Xiu}}
\author[1,4,5]{\newline\mbox{Ziqi Cai}}
\author[1,2]{\mbox{Zhen Li}}
\author[5]{\mbox{Erwin Wu}}
\author[1,\dagger]{\mbox{Kaipeng Zhang}}
\affiliation[1]{Alaya Lab}
\affiliation[2]{The University of Tokyo}
\affiliation[3]{\mbox{Georgia Institute of Technology}\newline}
\affiliation[4]{Peking University}
\affiliation[5]{\mbox{Institute of Science Tokyo}}

\input{sections/abstract}

\github{\url{https://zixiangao.github.io/StreamDuo/}}
\Code{\url{https://github.com/ZixianGao/StreamDuo-code}}
\correspondence{Kaipeng Zhang}
\date{\today}
\begin{document}
\maketitle

\vspace{10pt}
\begingroup
    \centering
    \includegraphics[width=\linewidth]{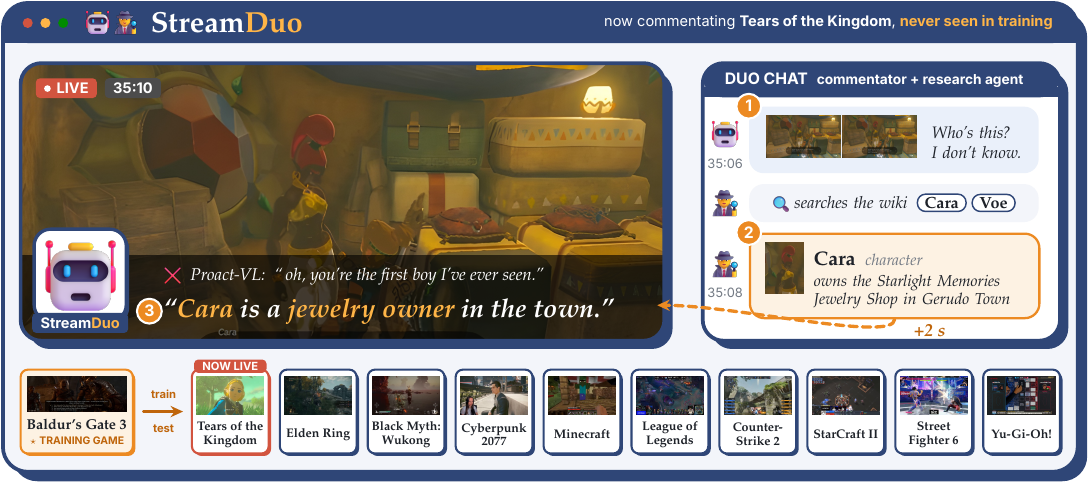}\vspace{5pt}
    \captionsetup{hypcap=false}
    \captionof{figure}{StreamDuo commentates Tears of the Kingdom, a game it has never seen, after training on
    Baldur's Gate~3 alone. When the commentator cannot name a character, it asks the research agent with the
    frames, and the wiki entry that arrives two seconds later lets it say who she is. Proact-VL, retrained on the same data, says something unrelated.}
    \label{fig:teaser}
\endgroup

\input{sections/introduction}
\input{sections/related_work}
\input{sections/method}
\input{sections/experiments}
\input{sections/conclusion}

\bibliographystyle{plainnat}
\bibliography{references}

\beginappendix
\input{sections/appendix}

\end{document}

%% file: sections/abstract.tex
\abstract{
Streaming video LLMs can now commentate live gameplay, but they are trained and tested on the same games. We
study cross-game streaming commentary, in which a model trained on one game commentates games it has never seen,
and find that current commentators transfer their timing but not their knowledge of the game world. We propose
StreamDuo, which pairs a trainable commentator with a frozen research agent that reads the wiki of the game being
played. When the commentator sees something it cannot name, it sends the current frames, and the research agent
returns a few wiki entries two seconds later while the stream continues. We train only the commentator, on a
single game, in two stages. Information-gain labels teach it to ask only where an answer would help it say what
the human caster said, and on-policy self-distillation from a copy of itself that sees the ideal answer teaches
it to use noisy answers. Trained on Baldur's Gate~3 and tested on ten unseen games, StreamDuo raises entity recall
from 0.082 to 0.116 over Proact-VL trained on the same data, cuts source-game leakage from 9.9\% to 1.0\%, and
achieves the best timing and the highest judged liveliness and overall quality among streaming models.
}

%% file: sections/introduction.tex
\section{Introduction}
\label{sec:intro}

Live game streaming has become one of the largest forms of online video, and commentary is what turns raw
gameplay into a show. A good caster reacts the moment something happens and tells the audience who just
appeared, what item was picked up and where the fight is taking place. Automating this requires a video
model that watches an unbounded stream, decides every second whether to speak, and says something both
timely and correct about the game world. Video LLMs~\cite{lin2024video,li2024llava,wang2025internvl3,bai2025qwen3}
understand short clips well, and streaming video
LLMs~\cite{chen2024videollm,chen2025live,wang2025videollm,zhang2024flash,qian2025dispider,xu2026streamingvlm,wang2026mmduet2}
decide frame by frame whether to respond. Automatic commentary was first studied for sports and
esports~\cite{mkhallati2023soccernet,rao2024matchtime,tanaka2021lol,zhang2022moba}, where the teams and rules
are known in advance. Proact-VL~\cite{yan2026proact} recently brought streaming video LLMs to game commentary
and reports strong timing and quality on a multi-game benchmark.

These systems are trained and evaluated on the same games. In practice a commentator constantly meets
titles, characters and items it has never seen, and the entities that make commentary informative are
exactly the ones that change from one game to the next. We therefore study cross-game streaming
commentary, in which a model is trained on a single game and commentates games that never appear in
training. Under this setting the standard recipe breaks down in a revealing way (Fig.~\ref{fig:teaser}).
Retrained on Baldur's Gate~3 alone, a Proact-VL commentator rarely names what is on screen in ten unseen
games, and when it does, it often reaches for its training game, for example calling a character in Tears
of the Kingdom ``Karlach''. Timing transfers across games, but knowledge does not.

A natural remedy is to give the model the missing knowledge. Retrieval-augmented
generation~\cite{lewis2020retrieval,asai2024self,jiang2023active,luo2026video}, tool-using
agents~\cite{schick2023toolformer,yao2022react,jin2025search} and multi-agent systems~\cite{wu2023autogen} all
supply external information to a model. Live commentary, however, imposes constraints these methods do not
address. Knowledge retrieved before the stream starts is mostly irrelevant to the current second, and in our
experiments a game-level dossier barely changes what the commentator says. Retrieving at every second floods
the context. Text queries are a further bottleneck, since the commentator would have to describe in words
something it cannot recognize. Most importantly, the commentator must decide when an answer is worth asking
for, which depends on what this particular model does not know. Language models can partly tell what they do
not know~\cite{kadavath2022language,wang2023self,jeong2024adaptive}, but streaming video models have no such
mechanism.

We propose StreamDuo, a two-agent streaming commentator. A commentator agent watches the stream and talks.
When it sees something it cannot name, it asks a research agent by sending the current frames. The
research agent, a frozen vision-language model with access to the game's wiki, returns a few entries two
seconds later while the stream continues. Perception and language, which transfer across games, stay in the
commentator, and game-specific knowledge is looked up at test time in whatever game is being played. We
train only the commentator, and only on the source game, in two stages. The first teaches it when to ask,
with ask labels placed where an answer would actually help the model reproduce what the human caster said.
The second teaches it how to use noisy answers, through on-policy self-distillation~\cite{zhao2026self} from
a privileged copy of itself that sees an ideal answer containing the right entry.
On ten unseen games, StreamDuo reduces source-game leakage from 9.9\% to 1.0\% and raises entity recall
from 0.082 to 0.116 compared with Proact-VL trained on the same data, while also improving timing and the commentary quality judged by an LLM on all ten test games.

In summary, our contributions are as follows.
\begin{itemize}
  \item We formulate cross-game streaming commentary, together with a clean evaluation protocol and a
        source-game leakage metric, and show that current streaming commentators transfer their timing but not their knowledge of the game world.
  \item We propose StreamDuo, a two-agent commentator that asks a frozen research agent with frames and
        receives its answers while the stream continues, trained with information-gain ask labels and
        on-policy self-distillation from a privileged copy of itself.
  \item On ten unseen games, StreamDuo outperforms zero-shot streaming models, Proact-VL retrained on the
        same data, and the same models given the research agent's knowledge without interaction.
\end{itemize}

%% file: sections/related_work.tex
\section{Related Work}
\label{sec:related}

\paragraph{Streaming video LLMs and live commentary.} Video LLMs understand complete
clips~\cite{lin2024video,li2024llama,xu2024slowfast,li2024llava,wang2025internvl3,bai2025qwen3}, and streaming
variants process frames as they arrive, either by managing memory over long
streams~\cite{song2024moviechat,he2024ma,zhang2024flash,qian2024streaming,di2025streaming,xu2026streamingvlm}
or by deciding when to respond without being prompted~\cite{chen2024videollm,wang2025videollm,wang2026mmduet2,
qian2025dispider,wang2026streambridge,yang2026livestar}, as measured by online
benchmarks~\cite{lin2026streamingbench,niu2025ovo}. LiveCC~\cite{chen2025live} and
Proact-VL~\cite{yan2026proact} bring this to live commentary. Earlier commentary systems targeted a single
domain such as chess~\cite{jhamtani2018learning}, soccer~\cite{mkhallati2023soccernet,qi2023goal,rao2024matchtime},
racing~\cite{ishigaki2021generating} or esports~\cite{tanaka2021lol,zhang2022moba,zhang2024game}, often with
structured game data. All of these models speak from what is in their weights and context, and are evaluated
on the domains they are trained on. We test on games whose entities are never seen in training and let the
commentator obtain the missing knowledge while the stream runs.

\paragraph{Retrieval, tools and multiple agents.} Retrieval-augmented generation supplies external
text~\cite{lewis2020retrieval}, and adaptive variants retrieve only when the model is
uncertain~\cite{jiang2023active,asai2024self,jeong2024adaptive,wang2023self,kadavath2022language}. Knowledge-based visual
question answering retrieves encyclopedic entries for an image~\cite{marino2019ok,caffagni2024wiki,yan2024echosight}, and video RAG
retrieves auxiliary text for video understanding~\cite{luo2026video}. Tool-using agents call search while
reasoning~\cite{schick2023toolformer,yao2022react,jin2025search}, and multi-agent frameworks let models with different roles
cooperate~\cite{wu2023autogen}. In StreamDuo the query is the frames themselves, the answer arrives after a
delay while the commentator keeps talking, and the decision to ask is learned from how much an answer would
change this particular model's output.

\paragraph{Distillation with privileged information.} A teacher with access to information unavailable at test
time can guide a student~\cite{vapnik2009new,lopez2015unifying}, and on-policy distillation trains the student on its own
samples~\cite{agarwal2024policy,gu2024minillm} to avoid the mismatch between training on reference sequences and generating at test
time~\cite{bengio2015scheduled,ross2011reduction}. On-policy self-distillation~\cite{zhao2026self} uses the same model,
conditioned on privileged information, as its own teacher. We apply it to a two-agent setting in which the
privileged information is the answer the research agent should have returned, and sharpen the teacher's
use of that answer with context-aware decoding~\cite{shi2024trusting}.

%% file: sections/method.tex
\begin{figure*}[t]
    \centering
    \includegraphics[width=\linewidth]{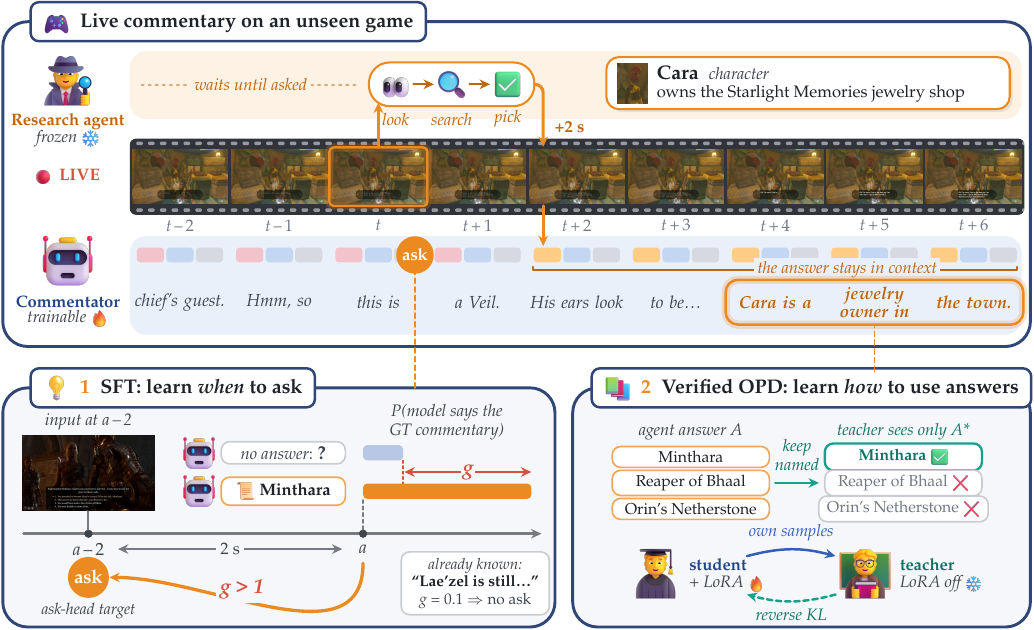}
    \caption{Overview of StreamDuo. Top: on an unseen game, the commentator asks the frozen research agent with
    the current frames, and the wiki entry it returns two seconds later lets the commentator name what is on
    screen. Bottom left: Stage 1 labels an ask only where an answer would make the human line more likely.
    Bottom right: Stage 2 distills a frozen copy of the commentator that sees the ideal answer into the
    student that sees the noisy one.}
    \label{fig:framework}
\end{figure*}

\section{Method}
\label{sec:method}

Our goal is a commentator that keeps talking at the right moments in a game it has never seen, and that says
the right names when it does. We first describe the streaming setting and why it fails across games
(Sec.~\ref{sec:setting}). We then present StreamDuo, which pairs the commentator with a research agent
(Sec.~\ref{sec:duo}), and the two training stages that teach the commentator when to ask
(Sec.~\ref{sec:when}) and how to use what it is told (Sec.~\ref{sec:how}). Fig.~\ref{fig:framework}
illustrates the whole pipeline.

\subsection{Streaming commentary across games}
\label{sec:setting}

A live stream arrives one second at a time. At second $t$ the commentator receives the frames $v_t$ of
that second and a short history $h_t$ made of the co-commentator's last line and its own recent lines.
Following Proact-VL~\cite{yan2026proact}, each second is one turn of a video LLM $\pi_\theta$. The turn ends with
a special flag token whose hidden state $z_t$ feeds a response head, which outputs the probability
$p_t = \sigma(w_s^{\top} z_t)$ that the commentator should speak now. If it speaks, the language model
generates a short line $y_t \sim \pi_\theta(\cdot \mid h_{\le t}, v_{\le t})$, and otherwise it stays
silent. Both parts are trained on human commentary with a language-modeling loss on the spoken lines and a
binary loss on the speak decisions,
\begin{equation}
  \mathcal{L}_{\text{base}} = \mathcal{L}_{\text{LM}} + \mathcal{L}_{\text{spk}} .
\end{equation}

In cross-game commentary, $\pi_\theta$ is trained on a single source game $G_s$ and evaluated on unseen
games $G_t \neq G_s$. The two parts of the model behave very differently under this shift. When to speak
depends on what happens on screen, such as a fight starting or a menu opening, and transfers well. What
to say about the entities on screen depends on knowledge of the game world, and a model trained on one game
only knows that one. Faced with an unfamiliar character, it either describes it vaguely or reaches for a
name it learned from the source game. To keep the model from tying names to what it memorized, we
rename the entities of the source game consistently and at random during training, so that a name in the
training data is evidence only within its own clip. This removes the shortcut but cannot supply the names
of a new game. That knowledge has to come from somewhere else at test time.

\subsection{StreamDuo}
\label{sec:duo}

StreamDuo splits the job between two agents with complementary strengths. The commentator agent
$\pi_\theta$ watches every second of the stream and decides when to talk, but only knows the game it was
trained on. The research agent $\mathcal{R}$ sees nothing of the stream unless asked, but can read the wiki
of whatever game is being played. Commentary emerges from their exchange. The commentator notices that it
does not recognize something, the research agent identifies it, and the commentator turns the answer into a
line.

\paragraph{The commentator asks with frames.} A natural interface would be a text query, but a model that
cannot recognize a character also cannot describe it well enough to retrieve it. The commentator therefore
sends the frames themselves. A second head on the same flag state estimates whether an answer is needed,
\begin{equation}
  q_t = \sigma(w_a^{\top} z_t),
\end{equation}
and the commentator sends $v_t$ to the research agent when $q_t$ exceeds a threshold.

\paragraph{The research agent looks and searches.} The research agent is a frozen vision-language model with
a search tool over the game's wiki. Before the stream starts, it writes a short dossier of the game, which
the commentator receives as general background. The dossier cannot say which of its entities is on screen
at a given second, so the real work happens when a question arrives. The research agent looks at the frames
it was sent, proposes a few search keys from what it sees, including any names readable on screen, searches
the wiki, looks at the frames again to check the candidates, and returns up to three entries, each with a
name, a type and a one-line description. We write the returned answer as $A_t$. Because the research agent
reads the frames rather than a description of them, it can identify a character the commentator could never have described in a text query.

\paragraph{Answers arrive while the stream continues.} Looking things up takes time, and the stream does
not wait. The answer to a question asked at second $t$ arrives $\delta$ seconds later and is placed in the
history $h_{t+\delta}$, where it stays available for the following seconds. Meanwhile the commentator keeps
deciding every second whether to speak. This asynchrony is part of the problem the commentator has to learn
to live with. It must ask before it needs the answer, and connect an answer to what it is seeing when the
answer arrives.

\paragraph{Speaking and stopping.} An answer often gives the commentator something to say, and a model that
has started an explanation tends to keep going after the moment passes. We therefore use a higher bar for
continuing than for starting. The commentator starts speaking when $p_t$ exceeds $\tau$, and after a run of
spoken seconds continues only when $p_t$ exceeds a larger threshold $\tau_c$. It can thus react immediately
to new events and stop once the event ends.

Only the commentator is trained, and only on the source game. The research agent stays frozen and brings
its knowledge from the wiki of the game at hand, so the same pair can be pointed at any game for which a
wiki exists. The commentator has to learn two things that
standard training does not teach, namely when an answer is worth asking for and how to turn a noisy answer
into the right line.

\subsection{Learning when to ask}
\label{sec:when}

The obvious supervision for the ask head is to ask shortly before the human caster names an entity. This
ignores the one thing that makes asking useful, which is whether the model already knows the answer. A
model that reliably recognizes the protagonist gains nothing from asking about her, while a model that
cannot tell two bosses apart gains a lot. The label should depend on the model.

We define the value of asking through the model itself. Let $y_a$ be the line the human says at second $a$
and let $A^{\star}_a$ be an ideal answer that contains the entity named in $y_a$. A frozen copy of the
commentator, $\pi_{\theta_0}$, scores this line twice, once with the answer in its history and once
without it. The information gain
\begin{equation}
  g_a = \log \pi_{\theta_0}\!\left(y_a \mid h_{\le a}, A^{\star}_a, v_{\le a}\right)
      - \log \pi_{\theta_0}\!\left(y_a \mid h_{\le a}, v_{\le a}\right)
\end{equation}
measures how much the answer helps the model say what the caster said. We label second $a - \delta$ as an
ask moment when $g_a > \gamma$, so that the answer would arrive in time. Lines the model can already
produce receive no label because their gain is small, and so do lines that no answer could help with, such
as tactical remarks. What remains are the moments when the model does not know and an answer would fix it.

The ask head is trained with a binary loss on these labels together with the base objective,
\begin{equation}
  \mathcal{L}_{1} = \mathcal{L}_{\text{LM}} + \mathcal{L}_{\text{spk}} + \lambda\,\mathcal{L}_{\text{ask}} .
\end{equation}
In the same stage the commentator learns to read answers. During training, each answer placed in the
history contains the entity the caster names next together with distractor entries from the same game,
and some answers miss the entity altogether, mimicking what a real research agent returns. Because entity
names are randomly renamed in each clip, the model cannot have memorized them and must take them from the answer that it receives.

\subsection{Learning to use the answer}
\label{sec:how}

Supervised training on answers teaches the format of an answer but not how to use it. The answers the
commentator receives at test time come from a real research agent. They are often plausible but wrong, and
the right entry, when present, is mixed with others. A model fine-tuned to copy entries from its history
will repeat whatever arrives, and a model that ignores its history will never say the new names. What we
want is a model that uses an entry when it matches the frames and otherwise does not.

We obtain this behavior with on-policy self-distillation~\cite{zhao2026self}. The student is the commentator
$\pi_\theta$, which sees a noisy answer $A$ exactly as it would at test time. The teacher is a frozen copy
of the same model, $\pi_{\theta_0}$, which sees the ideal answer $A^{\star}$ for the same moment. The
teacher is not a larger model. It knows more only because it is told which entry is right, and its behavior shows what using that entry looks like.

The effect of an answer on a fine-tuned model is often small, because the model has learned to rely on
the frames and its own memory. We make the teacher's use of the answer explicit by contrasting its
predictions with and without the answer~\cite{shi2024trusting},
\begin{equation}
  \log p_T(\cdot) = \log\operatorname{softmax}\!\Big( (1+\alpha)\, \ell_{\theta_0}(\cdot \mid A^{\star})
                    - \alpha\, \ell_{\theta_0}(\cdot \mid \varnothing) \Big),
\end{equation}
where $\ell_{\theta_0}$ denotes the teacher's next-token logits given the same frames and history and
$\alpha \ge 0$ controls the amplification. Tokens that become likely because of the answer, such as the
name of the entity on screen, are pushed up, and tokens the model would say anyway are left unchanged.

Training is on-policy. For a moment right after an answer arrives, the student samples its own lines
$\hat y \sim \pi_\theta(\cdot \mid h, A, v)$, and the teacher scores each of their tokens. We minimize the
reverse Kullback-Leibler divergence over the student's samples,
\begin{equation}
  \mathcal{L}_{2} = \mathbb{E}_{\hat y \sim \pi_\theta}\Big[ \sum_{k}
     \mathrm{KL}\big( \pi_\theta(\cdot \mid \hat y_{<k}, h, A, v) \,\big\|\, p_T(\cdot \mid \hat y_{<k}) \big) \Big].
\end{equation}
Sampling from the student matters here. The student is corrected on the lines it would actually say,
including the ones where it repeats a wrong entry, rather than on human lines it would never produce.
The reverse divergence makes the student commit to what the teacher considers likely instead of spreading
its probability over every name in the answer. Only a small set of low-rank adapter weights is updated in
this stage, so the speaking behavior learned before is preserved, and the teacher is simply the model with the LoRA adapter switched off during its forward pass.

At test time neither the teacher nor the ideal answer is used. The commentator watches the stream, asks with
its frames when its ask head fires, and speaks with whatever the research agent returns.

%% file: sections/experiments.tex
\section{Experiments}
\label{sec:exp}

\subsection{Setup}
\label{sec:setup}

\paragraph{Data.} We use the game commentary benchmark of Proact-VL~\cite{yan2026proact}, training on 10{,}286
Baldur's Gate~3 clips and testing on 2{,}470 clips from ten other games, from open-world adventures such as Tears
of the Kingdom to esports such as League of Legends. The original inputs leak game knowledge through the caster
persona, the earlier transcript and the other caster's lines. We therefore keep only the persona's speaking
style, remove earlier transcripts and replace game-specific names in the other caster's lines with typed
placeholders such as [PLAYER]. Game knowledge can then only come from the wiki of 190{,}722 pages.

\paragraph{Metrics.} For timing we report TimeDiff and F1 of Proact-VL. For knowledge we match wiki entities in
the two commentaries. Ent-R is the share of the human's entities that the model also names, Ent-F1 adds
precision, and K-strict counts only lines that need game knowledge. These count the first 126 words of each clip,
twice the human average, so that talking more cannot raise recall. Qwen3-30B~\cite{bai2025qwen3} scores PAUC, CC,
LiveU and FinalQ of Proact-VL on all test clips.

\paragraph{Baselines.} We compare with streaming video LLMs that decide when to respond,
VideoLLM-online~\cite{chen2024videollm}, LiveCC~\cite{chen2025live}, Dispider~\cite{qian2025dispider},
MMDuet~\cite{wang2025videollm}, StreamingVLM~\cite{xu2026streamingvlm}, MMDuet2~\cite{wang2026mmduet2} and
Proact-VL~\cite{yan2026proact}, each with its official code and the game title as context. We test them zero-shot
and after one epoch on the same Baldur's Gate~3 clips with the official recipe. Proact-VL has seen the test games,
so we retrain it from LiveCC-7B-Base. Dispider releases no training code, and for MMDuet2 we rerun only the
supervised stage. Each trained baseline is also given the research agent's dossier, to separate the knowledge itself from the way in which StreamDuo obtains it during the stream.

\paragraph{Implementation.} The commentator starts from LiveCC-7B-Base~\cite{chen2025live}, as Proact-VL does.
Stage~1 fine-tunes all parameters for one epoch at a learning rate of $10^{-5}$ with $\gamma = 1$ nat. Stage~2
trains a rank-32 LoRA~\cite{hu2021lora} for 300 steps with $\alpha = 1$. The research agent is
Qwen3-VL-8B~\cite{bai2025qwen3} with BM25 search~\cite{robertson2009probabilistic} and returns up to three
entries after $\delta = 2$ s. At test time it speaks when $p_t > 0.27$ and asks when $q_t > 0.9$. All runs use eight NVIDIA H100 GPUs for training and evaluation.

\input{sections/tables/main}

\subsection{Main results}
\label{sec:main}

Tab.~\ref{tab:main} shows that existing streaming models fail on unseen games in two ways. Zero-shot, they rarely speak
at the right moments. LiveCC names the most entities among them but has a TimeDiff of 3.03 s, and the others either stay
silent or describe the scene. After training on Baldur's Gate~3, Proact-VL and StreamingVLM learn when to speak, but their
entity recall stays below that of zero-shot LiveCC, and VideoLLM-online speaks more often without naming a single correct
entity. Models that decide silence through their language output, such as LiveCC and MMDuet2, collapse to a few words per
clip and mostly read the dialogue shown on screen. Timing transfers across games, knowledge does not. Adding the research
agent's game dossier to their context barely helps, for example raising the K-strict of Proact-VL by only 0.003, since a
dossier covers the whole game but cannot tell which entity is on screen. StreamDuo is best on all metrics except CC. Compared with Proact-VL, trained from the same base on the same clips, it
raises entity recall from 0.082 to 0.116 and K-strict from 0.085 to 0.127, while responding faster and receiving higher
judge scores. It also names more entities than zero-shot LiveCC without giving up timing. CC favors near-silent models in
this comparison, such as MMDuet2 with a few words per clip, and does not reflect the quality of the commentary itself.

\subsection{Contribution of each component}
\label{sec:abl_components}

\input{sections/tables/components}

Tab.~\ref{tab:abl_components} adds the components of StreamDuo one at a time, all with the same inference
settings. The first row follows the recipe of Proact-VL. Renaming the entities of the source game and adding the
dossier do not raise entity recall (0.082 to 0.083), which agrees with the dossier results in
Tab.~\ref{tab:main}, but they make the commentary more natural, raising LiveU from 4.91 to 5.03. Asking the
research agent with the frames adds what the dossier cannot provide, the identity of what is on screen, and lifts
Ent-R to 0.092 while also improving timing. OPSD gives the largest gain of the three components. It raises Ent-R from 0.092 to 0.116 and
K-strict from 0.094 to 0.127, as the commentator learns to use the answers it receives instead of only asking
for them. Together the three components raise Ent-R by 0.034 and K-strict by 0.042 over the Proact-VL recipe,
and the full model is best on every metric.

\subsection{When to ask}
\label{sec:abl_ask}

\input{sections/tables/ask}

The upper part of Tab.~\ref{tab:abl_ask} changes the labels that train the ask head. Labeling every second in
which the human names an entity also marks names on screen or already known, and every metric drops,
Ent-R from 0.092 to 0.079 and FinalQ from 3.48 to 3.38. The lower part changes the asking policy of the full
model at test time. Without asking, timing and judge scores stay the same but knowledge falls, to an Ent-R of
0.104 and a K-strict of 0.114. Asking every 3 s nearly doubles recall, but many of the added names do not fit
the moment, so Ent-F1 falls to 0.105 and all judge scores drop. The learned policy asks 264 times over 2{,}470
clips and gives the best balance, with the highest Ent-F1 of 0.141.

\subsection{When does the commentator ask?}
\label{sec:ask_analysis}

\begin{wrapfigure}[10]{r}{0.40\textwidth}
    \centering
    \includegraphics[width=\linewidth]{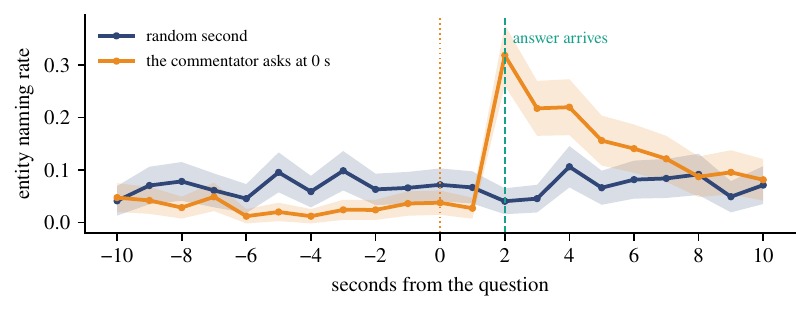}
    \caption{Entity naming around questions (orange) and random seconds (blue). The answer arrives at 2 s.}
    \label{fig:ask_named}
\end{wrapfigure}

Fig.~\ref{fig:ask_named} aligns the commentary of StreamDuo with its own questions on the test games. Questions
are rare, 264 over 2{,}470 clips. Before a question, the commentator names an entity in only 1 to 4\% of the
seconds, below the 7\% of a random second, so it asks when it has no name to offer. When the answer arrives two
seconds later, the share jumps to 32\%, 4.6 times that of a random second. It stays above the random rate for five more seconds, at 22\% one and two seconds after the answer and 12\% five seconds after it, and returns to the base rate within about eight seconds. The jump comes from the answer and not from the scene. With asking disabled, the same
seconds name an entity only 5\% of the time. The commentator thus uses an answer at once.

\needspace{9\baselineskip}
\subsection{What Stage~2 needs}
\label{sec:abl_stage2}

\begin{wrapfigure}[17]{r}{0.34\textwidth}
    \centering
    \includegraphics[width=\linewidth]{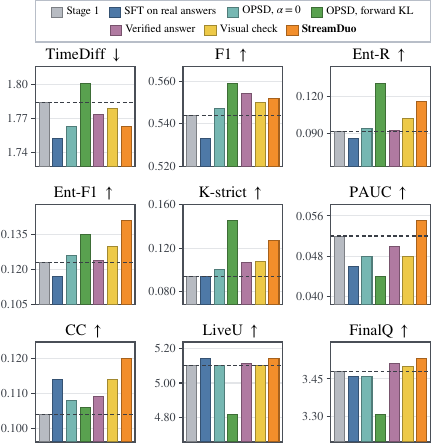}
    \caption{Stage~2 variants. Dashed line: Stage~1.}
    \label{fig:abl_stage2}
\end{wrapfigure}

Fig.~\ref{fig:abl_stage2} varies the loss and the teacher of Stage~2. Fine-tuning on the real answers of the
research agent gives the lowest TimeDiff and a higher LiveU, but entity recall falls below that of Stage~1 (0.086
against 0.092). Without amplification ($\alpha = 0$) the gain over Stage~1 is small. The forward divergence spreads
the student over all the tokens the teacher allows, so the student names an entity wherever the amplified teacher
gives it some probability. It reaches the highest Ent-R (0.131) and K-strict (0.145), but one in five of its entity
mentions is a name from Baldur's Gate~3 (0.193, against 0.010 for the full model), and its LiveU and FinalQ fall to
4.82 and 3.31, the lowest of all variants. All three loss variants also lower FinalQ. Weaker teachers keep the
quality of the commentary but add less knowledge. A teacher that only knows which received entries the human names,
or one that checks them on screen with a frozen VLM, reaches an Ent-R of 0.093 or 0.102. Only the full recipe, the
reverse divergence with amplification and the ideal answer, improves all knowledge and judge metrics over Stage~1,
raising Ent-R to 0.116 and K-strict to 0.127 without leaking names from the source game.

\subsection{Swapping the research agent}
\label{sec:agent_swap}

\input{sections/tables/agent}

Tab.~\ref{tab:agent_swap} replaces the research agent at test time with other Qwen3-VL models, from 2B to
30B-A3B, keeping the commentator and the wiki search unchanged. All agents answer every second of the test set,
so the 8B row equals the main result. The 8B agent gives the most knowledge, but the choice of agent matters
little. Ent-R stays between 0.101 and 0.116 and the judge scores within 0.03 of each other, even though the 2B
agent answers only 18{,}103 of the 80{,}529 moments against about 50{,}000 for the larger ones, and 30B-A3B does
not improve on 8B. Within this family and search, the size of the research agent is not the bottleneck. What
matters more is how the commentator uses the entries it receives. With the same 8B agent, Stage~2 alone raises
Ent-R from 0.092 to 0.116 (Sec.~\ref{sec:abl_stage2}), far more than any change of agent.


\subsection{Qualitative examples}
\label{sec:qualitative}

\begin{figure}[t]
    \centering
    \includegraphics[width=\linewidth]{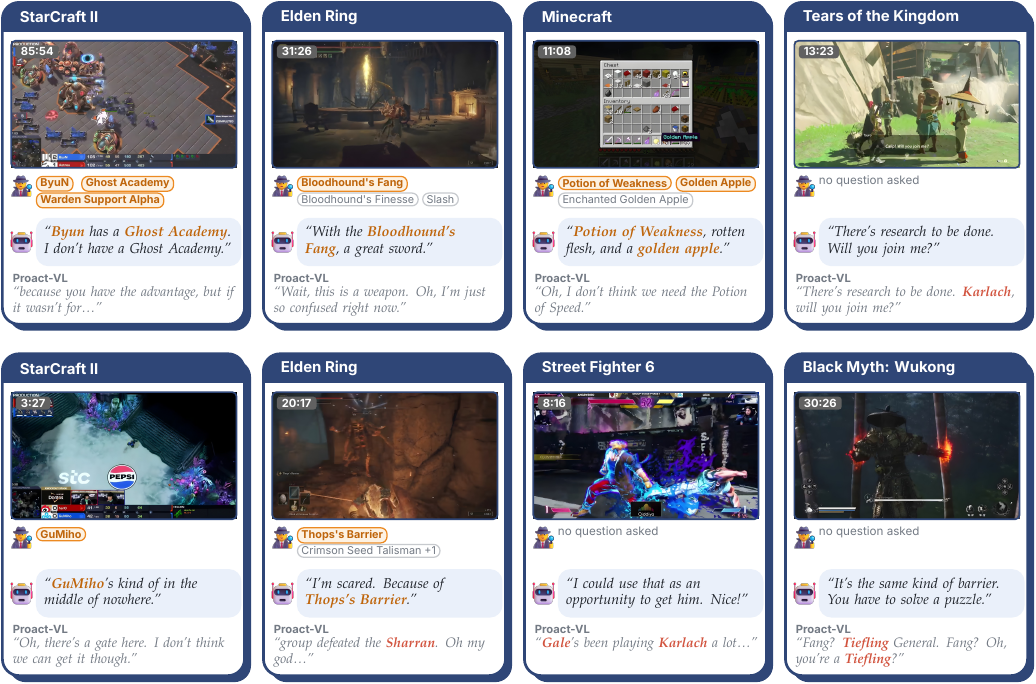}
    \caption{Examples on unseen games, with the frame at the second the research agent's answer arrives. Orange
entries are used by the commentator, and gray ones are not. Proact-VL is retrained on the same data and shown
at the same seconds, and names from Baldur's Gate~3 are in red.}
    \label{fig:qa}
\end{figure}

Fig.~\ref{fig:qa} shows StreamDuo and Proact-VL at the same seconds of seven unseen games. In the first five
examples the commentator asks and, two seconds later, names what the answer contains. In StarCraft~II it names
Byun's Ghost Academy, which the human caster also names, and the player GuMiho in another match. In Elden Ring
and Minecraft the entry is also on screen, in the equipment menu, the chest and the skill slot of Thops's
Barrier, so the research agent finds it from the frames although the commentator cannot read it. The answers
are not clean. Each holds up to three entries, and the commentator mostly keeps the ones that match the frames,
Bloodhound's Fang but not Bloodhound's Finesse or Slash, and the golden apple but not the enchanted one. It does
not always filter well, and in StarCraft~II it also repeats the Warden Support Alpha, an entry unrelated to the
match. Proact-VL speaks at the same seconds but stays vague, as in ``Wait, this is a weapon'', names the wrong
potion, or calls an Elden Ring enemy group the Sharran, a faction of Baldur's Gate~3. The last three examples
show what happens without a question. In Tears of the Kingdom, Street Fighter~6 and Black Myth: Wukong the
commentator does not ask, since its ask score stays below the threshold. StreamDuo describes the scene without a
name, while Proact-VL fills in names from its training game, addressing a character as Karlach, saying a
fighter has been playing Karlach, and calling a boss a Tiefling. Such names are not rare. Over all test clips,
names from the source game make up 9.9\% of the entity mentions of Proact-VL, 200 of its 2{,}011, and 1.0\% of
ours, 34 of 3{,}363, although StreamDuo names almost twice as many entities. Without an answer, the commentator
leaves a name out instead of borrowing one from the game it was trained on.

%% file: sections/tables/main.tex
\definecolor{oursbg}{HTML}{EAF2FB}
\newcommand{\best}[1]{\textbf{#1}}
\newcommand{\secbest}[1]{\underline{#1}}
\newcommand{\venue}[1]{{\scriptsize\color{gray}#1}}
\newcommand{\dossier}{\hspace{0.8em}{\color{gray}\itshape + agent dossier}}

\begin{table}[t]
\centering
\caption{\textbf{Cross-game streaming commentary.} Models are trained on Baldur's Gate~3 only (or not trained)
and evaluated on 2{,}470 clips from ten unseen games. Knowledge metrics use the first 126 words of each clip
(twice the human average), so a model cannot gain recall by talking more. Judge metrics are scored by Qwen3-30B.
$^\ddagger$Supervised stage of the official recipe only, without RL.
\best{Bold}: best; \secbest{underline}: second best.}
\label{tab:main}
\small
\setlength{\tabcolsep}{3.2pt}
\renewcommand{\arraystretch}{1.08}
\begin{tabular}{@{}l l cc ccc cccc@{}}
\toprule
& & \multicolumn{2}{c}{Timing} & \multicolumn{3}{c}{Knowledge} & \multicolumn{4}{c}{LLM judge} \\
\cmidrule(lr){3-4}\cmidrule(lr){5-7}\cmidrule(l){8-11}
Method & Venue & TimeDiff$\downarrow$ & F1$\uparrow$ & Ent-R$\uparrow$ & Ent-F1$\uparrow$ & K-strict$\uparrow$
       & PAUC$\uparrow$ & CC$\uparrow$ & LiveU$\uparrow$ & FinalQ$\uparrow$ \\
\midrule
\multicolumn{11}{@{}l}{\textit{Zero-shot}} \\
VideoLLM-online~\cite{chen2024videollm} & \venue{CVPR'24}  & 11.63 & 0.063 & 0.002 & 0.003 & 0.001 & 0.000 & 0.134 & 3.49 & 2.39 \\
LiveCC-7B-Instruct~\cite{chen2025live}  & \venue{CVPR'25}  & 3.03  & 0.470 & \secbest{0.099} & \secbest{0.139} & \secbest{0.115} & 0.031 & 0.127 & 3.56 & 2.59 \\
Dispider~\cite{qian2025dispider}        & \venue{CVPR'25}  & 6.10  & 0.156 & 0.043 & 0.042 & 0.055 & 0.006 & 0.083 & 3.61 & 2.78 \\
MMDuet~\cite{wang2025videollm}          & \venue{EMNLP'25} & 4.01  & 0.297 & 0.017 & 0.023 & 0.020 & 0.002 & 0.049 & 3.55 & 2.50 \\
StreamingVLM~\cite{xu2026streamingvlm}  & \venue{ICLR'26}  & \secbest{1.83} & 0.509 & 0.076 & 0.081 & 0.086 & 0.025 & 0.097 & 4.58 & 3.25 \\
MMDuet2~\cite{wang2026mmduet2}          & \venue{ICLR'26}  & 3.52  & 0.330 & 0.073 & 0.091 & 0.083 & 0.026 & 0.107 & \secbest{5.05} & \secbest{3.47} \\
\midrule
\multicolumn{11}{@{}l}{\textit{Trained on Baldur's Gate 3}} \\
VideoLLM-online~\cite{chen2024videollm} & \venue{CVPR'24}  & 5.41  & 0.228 & 0.000 & 0.000 & 0.000 & 0.001 & 0.124 & 3.71 & 2.27 \\
\dossier                                &                  & 4.25  & 0.231 & 0.001 & 0.001 & 0.001 & 0.001 & 0.129 & 3.63 & 1.98 \\
LiveCC-7B-Instruct~\cite{chen2025live}  & \venue{CVPR'25}  & 24.08 & 0.036 & 0.028 & 0.053 & 0.034 & 0.031 & 0.306 & 1.73 & 1.55 \\
\dossier                                &                  & 19.89 & 0.072 & 0.031 & 0.050 & 0.034 & 0.030 & 0.252 & 2.35 & 1.81 \\
MMDuet~\cite{wang2025videollm}          & \venue{EMNLP'25} & 3.62  & 0.256 & 0.008 & 0.015 & 0.009 & 0.001 & 0.094 & 4.30 & 2.39 \\
\dossier                                &                  & 16.46 & 0.072 & 0.004 & 0.009 & 0.010 & 0.001 & 0.166 & 4.17 & 2.57 \\
StreamingVLM~\cite{xu2026streamingvlm}  & \venue{ICLR'26}  & 1.95  & 0.478 & 0.076 & 0.111 & 0.083 & 0.039 & 0.090 & 4.77 & 3.20 \\
\dossier                                &                  & 2.07  & 0.467 & 0.066 & 0.097 & 0.071 & 0.036 & 0.087 & 4.73 & 3.25 \\
MMDuet2$^\ddagger$~\cite{wang2026mmduet2} & \venue{ICLR'26} & 24.07 & 0.016 & 0.021 & 0.039 & 0.018 & 0.029 & \secbest{0.312} & 1.75 & 1.52 \\
\dossier                                &                  & 24.33 & 0.015 & 0.020 & 0.038 & 0.018 & 0.027 & \best{0.313} & 1.71 & 1.50 \\
Proact-VL~\cite{yan2026proact}          & \venue{ICML'26}  & 1.84  & 0.536 & 0.082 & 0.118 & 0.085 & \secbest{0.050} & 0.095 & 4.91 & 3.32 \\
\dossier                                &                  & 1.88  & \secbest{0.548} & 0.083 & 0.111 & 0.088 & 0.048 & 0.089 & 4.80 & 3.30 \\
\midrule
\rowcolor{oursbg}
\textbf{StreamDuo (ours)}               & \venue{--}       & \best{1.76} & \best{0.552} & \best{0.116} & \best{0.141} & \best{0.127}
                                                           & \best{0.055} & 0.120 & \best{5.14} & \best{3.53} \\
\bottomrule
\end{tabular}
\end{table}

%% file: sections/tables/components.tex
\newcommand{\ablyes}{\checkmark}
\newcommand{\ablno}{{\color{gray!70}$\times$}}

\begin{table}[htb]
\centering
\caption{\textbf{Ablation of StreamDuo components.} Rename \& dossier: random entity renaming during training
plus the game dossier in context. Asking: Stage~1. OPSD: Stage~2. The first row corresponds to the Proact-VL
baseline trained on Baldur's Gate~3. Best results are in bold.}
\label{tab:abl_components}
\small
\setlength{\tabcolsep}{3.6pt}
\renewcommand{\arraystretch}{1.1}
\begin{tabular}{@{}ccc cc ccc cccc@{}}
\toprule
\multicolumn{3}{c}{Components} & \multicolumn{2}{c}{Timing} & \multicolumn{3}{c}{Knowledge} & \multicolumn{4}{c}{LLM judge} \\
\cmidrule(r){1-3}\cmidrule(lr){4-5}\cmidrule(lr){6-8}\cmidrule(l){9-12}
Rename \& dossier & Asking & OPSD & TimeDiff$\downarrow$ & F1$\uparrow$ & Ent-R$\uparrow$ & Ent-F1$\uparrow$ & K-strict$\uparrow$
       & PAUC$\uparrow$ & CC$\uparrow$ & LiveU$\uparrow$ & FinalQ$\uparrow$ \\
\midrule
\ablno  & \ablno  & \ablno  & 1.84 & 0.536 & 0.082 & 0.118 & 0.085 & 0.050 & 0.095 & 4.91 & 3.32 \\
\ablyes & \ablno  & \ablno  & 1.88 & 0.523 & 0.083 & 0.116 & 0.090 & 0.045 & 0.093 & 5.03 & 3.38 \\
\ablyes & \ablyes & \ablno  & 1.78 & 0.544 & 0.092 & 0.123 & 0.094 & 0.052 & 0.104 & 5.10 & 3.48 \\
\midrule
\rowcolor{oursbg}
\ablyes & \ablyes & \ablyes & \best{1.76} & \best{0.552} & \best{0.116} & \best{0.141} & \best{0.127}
                            & \best{0.055} & \best{0.120} & \best{5.14} & \best{3.53} \\
\bottomrule
\end{tabular}
\end{table}

%% file: sections/tables/ask.tex
\begin{table}[htb]
\centering
\caption{\textbf{When to ask.} \emph{Top:} Stage~1 ask labels, placed at every human entity mention (naive)
or only where the answer yields an information gain of $\gamma = 1$ nat (info.\ gain).
\emph{Bottom:} test-time asking policy of the full model. Shaded rows are the settings used by StreamDuo;
best results within each block are in bold.}
\label{tab:abl_ask}
\small
\setlength{\tabcolsep}{2pt}
\renewcommand{\arraystretch}{1.1}
\begin{tabularx}{\linewidth}{@{}l*{9}{>{\centering\arraybackslash}X}@{}}
\toprule
& \multicolumn{2}{c}{Timing} & \multicolumn{3}{c}{Knowledge} & \multicolumn{4}{c}{LLM judge} \\
\cmidrule(lr){2-3}\cmidrule(lr){4-6}\cmidrule(l){7-10}
Setting & TimeDiff$\downarrow$ & F1$\uparrow$ & Ent-R$\uparrow$ & Ent-F1$\uparrow$ & K-strict$\uparrow$
        & PAUC$\uparrow$ & CC$\uparrow$ & LiveU$\uparrow$ & FinalQ$\uparrow$ \\
\midrule
\multicolumn{10}{@{}l}{\textit{Stage~1 ask labels}} \\
Naive       & 1.903 & 0.534 & 0.079 & 0.107 & 0.087 & 0.046 & 0.098 & 5.01 & 3.38 \\
\rowcolor{oursbg}
Info.\ gain & \best{1.784} & \best{0.544} & \best{0.092} & \best{0.123} & \best{0.094}
            & \best{0.052} & \best{0.104} & \best{5.10} & \best{3.48} \\
\midrule
\multicolumn{10}{@{}l}{\textit{Test-time asking}} \\
Never       & 1.764 & 0.553 & 0.104 & 0.127 & 0.114 & 0.051 & 0.117 & \best{5.14} & \best{3.54} \\
Every 3\,s  & 1.800 & \best{0.563} & \best{0.199} & 0.105 & \best{0.208} & 0.043 & 0.086 & 4.76 & 3.36 \\
\rowcolor{oursbg}
Learned     & \best{1.763} & 0.552 & 0.116 & \best{0.141} & 0.127 & \best{0.055} & \best{0.120} & \best{5.14} & 3.53 \\
\bottomrule
\end{tabularx}
\end{table}

%% file: sections/tables/agent.tex
\begin{table}[htb]
\centering
\caption{\textbf{Swapping the research agent at test time} without retraining the commentator.
All agents answer the same moments, every second of the 2{,}470 test clips (80{,}529 in total).
Answers: number of these moments for which the agent returns an answer. The shaded row is the default
agent used by StreamDuo. Best results are in bold.}
\label{tab:agent_swap}
\small
\setlength{\tabcolsep}{2pt}
\renewcommand{\arraystretch}{1.1}
\begin{tabularx}{\linewidth}{@{}l c *{5}{>{\centering\arraybackslash}X}@{}}
\toprule
& & \multicolumn{3}{c}{Knowledge} & \multicolumn{2}{c}{LLM judge} \\
\cmidrule(lr){3-5}\cmidrule(l){6-7}
Research agent & Answers & Ent-R$\uparrow$ & Ent-F1$\uparrow$ & K-strict$\uparrow$ & LiveU$\uparrow$ & FinalQ$\uparrow$ \\
\midrule
Qwen3-VL-2B      & 18{,}103 & 0.101 & 0.125 & 0.107 & 5.14 & 3.53 \\
Qwen3-VL-4B      & 50{,}445 & 0.105 & 0.129 & 0.111 & 5.13 & 3.52 \\
\rowcolor{oursbg}
Qwen3-VL-8B      & 52{,}548 & \best{0.116} & \best{0.141} & \best{0.127} & 5.14 & 3.53 \\
Qwen3-VL-30B-A3B & 54{,}888 & 0.107 & 0.129 & 0.117 & \best{5.15} & \best{3.55} \\
\bottomrule
\end{tabularx}
\end{table}

%% file: sections/conclusion.tex
\section{Conclusion}
\label{sec:conclusion}

We studied cross-game streaming commentary, in which a model trained on one game commentates games it has
never seen, and found that current streaming commentators transfer their timing but not their knowledge.
To address this, StreamDuo pairs a trainable commentator with a frozen research agent that it asks with its
frames and that answers from the game's wiki while the stream continues. Information-gain labels teach the
commentator when an answer is worth asking for, and on-policy self-distillation from a privileged copy of
itself teaches it to use noisy answers. Trained on Baldur's Gate~3 alone, StreamDuo names more of what is on
screen in ten unseen games, rarely falls back on names from its training game, and keeps the timing and
quality of the strongest streaming commentators, without any training on the ten games it is tested on.

\section*{Acknowledgments}
This study was carried out using the TSUBAME4.0 supercomputer at Institute of Science Tokyo.

%% file: sections/appendix.tex
\section{Source-game leakage}
\label{sec:leakage}

\input{sections/tables/leakage}

A commentator trained on one game can fill an entity it does not know with a name it memorized from that game.
We call this source-game leakage and measure it as the share of Baldur's Gate~3 names among all entity names a
model says in the unseen games, counting only names that exist neither in the test game nor in its wiki.
Tab.~\ref{tab:leakage} reports it for all models trained on Baldur's Gate~3, on all 2{,}470 clips and without a
word cap.

The two models that speak the most entity names before StreamDuo leak the most. 17.3\% of the names said by
StreamingVLM and 9.9\% of those said by Proact-VL come from Baldur's Gate~3, such as Karlach in Street Fighter~6
or Lae'zel in League of Legends. Leakage is worst in esports, where the commentary is full of player and champion
names the model has never seen. In League of Legends, 54.6\% of the names said by Proact-VL come from Baldur's
Gate~3, against 7.8\% for StreamDuo. The agent dossier lowers the overall share only a little, to 14.7\% and
6.3\%, since it lists the entities of the game but cannot tell which one is on screen. The other baselines leak
less because they name few entities at all, with fewer than 900 test-game names over all 2{,}470 clips. A low
share then says little about their knowledge. StreamDuo names the most entities of the test games, 3{,}329, and
keeps leakage at 1.0\%.

Each part of StreamDuo contributes to this. Run with the same inference settings as StreamDuo, the recipe of
Proact-VL leaks 5.8\% of its names. Random renaming with the dossier in context (Stage~0) lowers this to 2.1\%,
since a name from the source game is no longer reliable evidence during training. Learning when to ask
(Stage~1) lowers it further to 0.8\%. With asking disabled at test time, the full model leaks 2.9\%, three times
as much as with asking. When the commentator does not know an entity, it asks the research agent instead of
reaching for a name from the source game. The loss of Stage~2 matters as well. With the forward instead of the
reverse divergence, leakage rises to 19.3\%, the highest of all models, because the student learns to name an
entity wherever the teacher gives it some probability (Sec.~\ref{sec:abl_stage2}).

\subsection{Efficiency}
\label{sec:latency}

\input{sections/tables/efficiency}

A live commentator has to keep up with the stream, and an answer is only useful if it arrives while the entity
is still on screen. Tab.~\ref{tab:latency} measures both agents on one H100 GPU with one request at a time, as
in a live stream, using the same models, prompts and settings as in all other experiments.

The research agent answers in 0.41~s on average and in at most 0.71~s over 200 test moments, so every answer is
ready before the 2~s at which StreamDuo places it in the history. The delay of 2~s therefore leaves a margin of
more than a second, and a slower or larger agent would still fit. Most of this time goes to proposing search
keys from the frames (0.32~s). Picking up to three entries from the retrieved candidates takes 0.08~s, and the
BM25 search over the wiki takes about 2~ms. Building the search index is done once before the stream and takes
1.5~s.

The commentator needs 0.26~s of compute for each second of video, about four times faster than real time, and
at most 0.33~s on the slowest of 20 test clips. This includes decoding the frames, the speak and ask heads and
the generation of each line. All 20 clips are mostly spoken, so the measurement covers the most expensive case.

The research agent runs in parallel and is called only when the commentator asks. In one test run, the
commentator asked 264 times over the 26~hours of test video, about once every six minutes, and in 243 of the
2{,}470 clips. The research agent is therefore idle most of the time, and its cost adds little to that of the
commentator. StreamDuo can commentate a live stream on two GPUs, one for each agent.

\section{More qualitative examples}
\label{app:qa}

\begin{figure}[p]
    \centering
    \includegraphics[width=\textwidth,height=0.93\textheight,keepaspectratio]{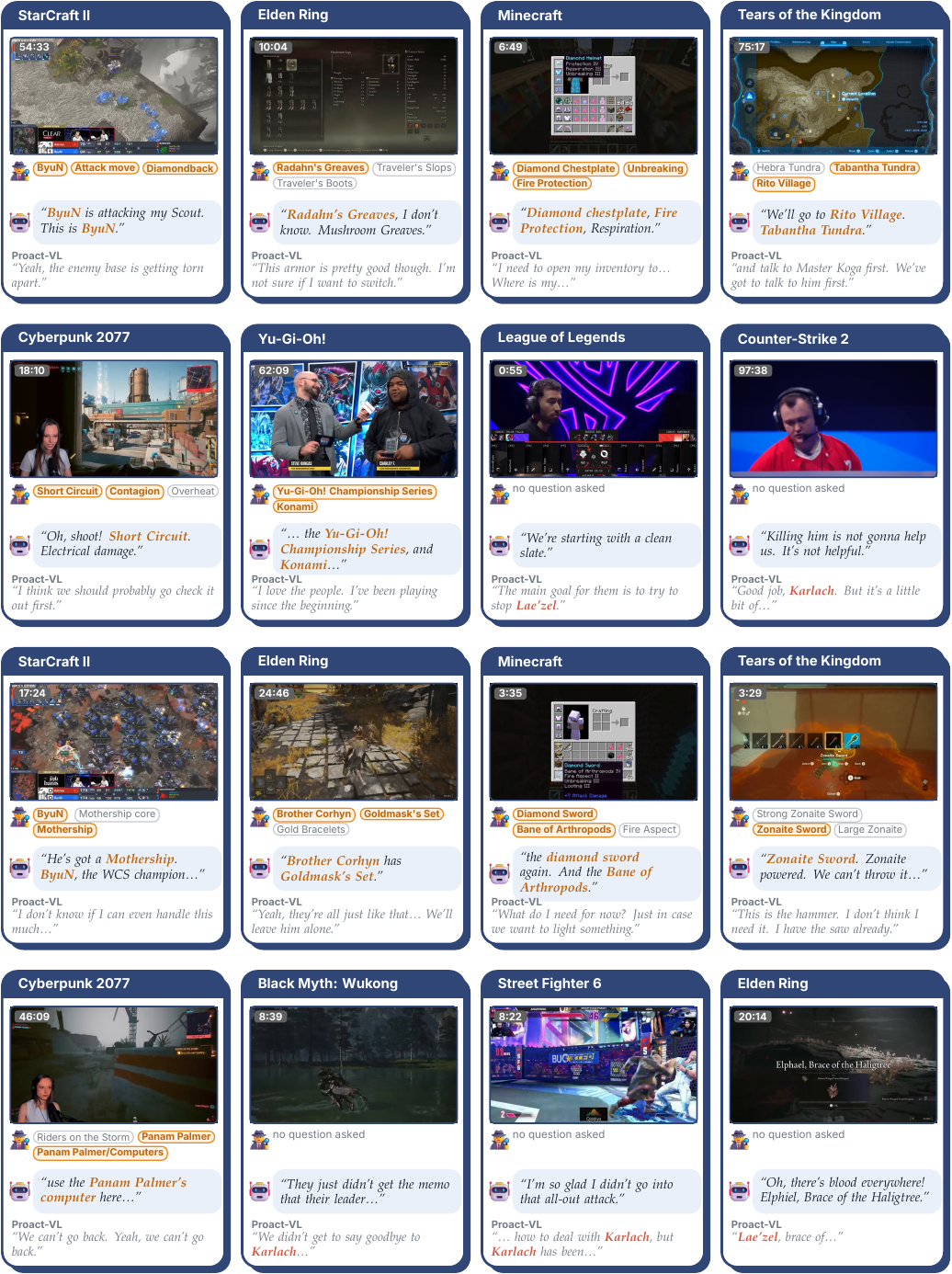}
    \caption{More examples from the ten unseen games, with StreamDuo and Proact-VL at the same seconds (1/2).
    Chips: the entries in the research agent's answer, in orange those the commentator names. Red: names from
    Baldur's Gate~3.}
    \label{fig:qa_app1}
\end{figure}

\begin{figure}[p]
    \centering
    \includegraphics[width=\textwidth,height=0.93\textheight,keepaspectratio]{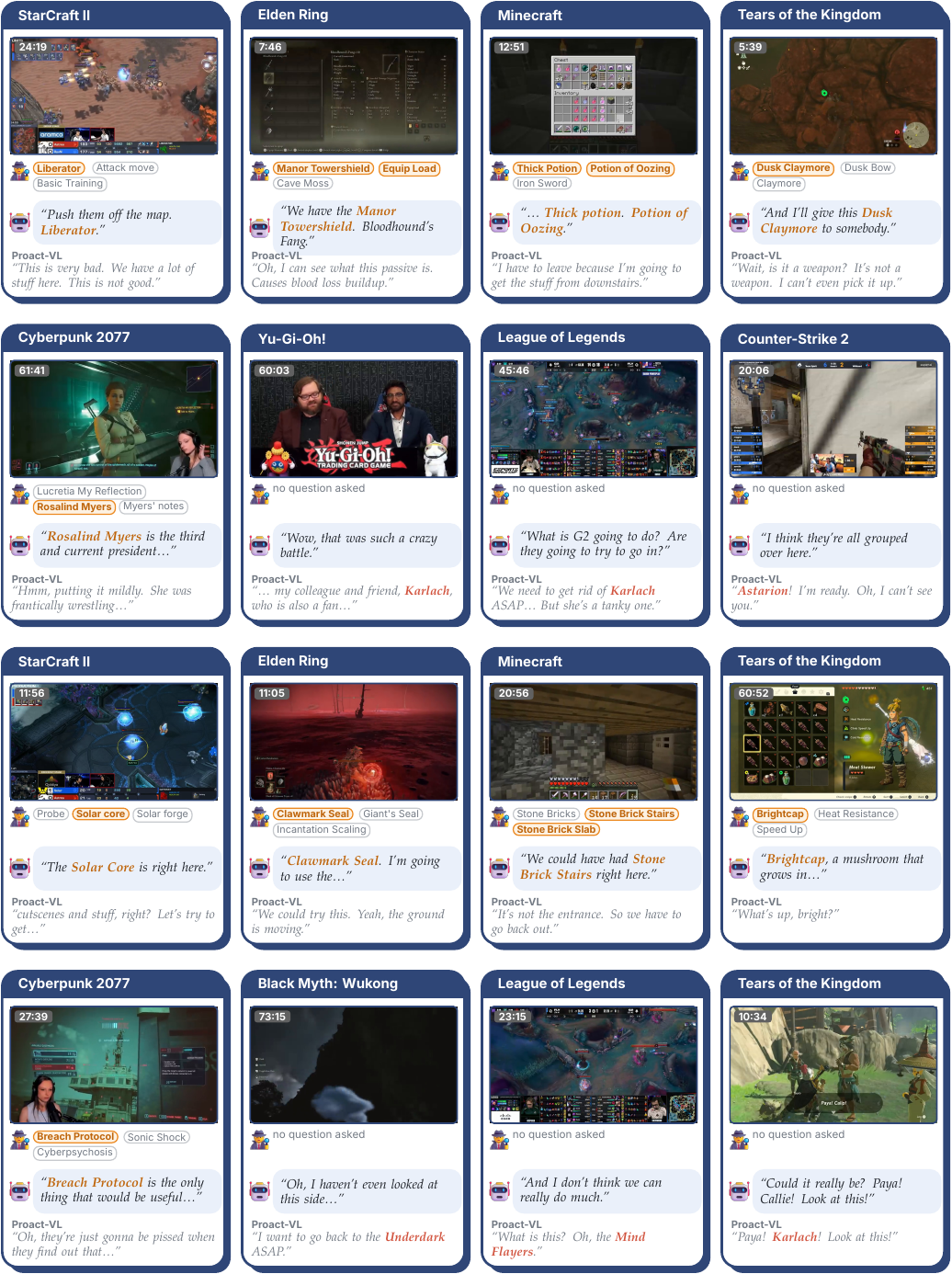}
    \caption{More examples from the ten unseen games (2/2).}
    \label{fig:qa_app2}
\end{figure}

Fig.~\ref{fig:qa_app1} and Fig.~\ref{fig:qa_app2} show 32 more examples from all ten unseen games, none of them
in Fig.~\ref{fig:qa}. Each card shows the frame at the given time, the entries that the research agent returns,
StreamDuo's commentary and that of Proact-VL, retrained on the same data, at the same seconds. All text is the
models' own output; ``\dots'' marks a cut.

In 21 examples the commentator asks and names what the answer contains within a few seconds. The entries cover
the main kinds of game knowledge: players and units in StarCraft~II (ByuN, Mothership, Liberator), items and
equipment in Elden Ring and Minecraft (Radahn's Greaves, Goldmask's Set, the enchantments of a diamond
chestplate), places in Tears of the Kingdom (Rito Village, Tabantha Tundra) and skills in Cyberpunk~2077 (Short
Circuit, Breach Protocol). The commentator does not repeat every entry. It keeps the Dusk Claymore but not the
Dusk Bow, and the Zonaite Sword but not the Strong Zonaite Sword. It sometimes reads the wiki entry instead of
commenting on the game, as in ``Rosalind Myers is the third and current president\dots'' or ``Brightcap, a
mushroom that grows in\dots''.

In the other 11 examples no question is asked. StreamDuo then describes the action or reads a name off the
screen, such as Elphael, Brace of the Haligtree in Elden Ring. At these seconds Proact-VL names characters,
creatures and places from Baldur's Gate~3: Lae'zel and the Mind Flayers in League of Legends, Karlach in
Counter-Strike~2, Street Fighter~6, Yu-Gi-Oh! and Tears of the Kingdom, and the Underdark in Black Myth: Wukong.
In the examples where StreamDuo asks, Proact-VL at the same seconds names none of the entries and stays vague, as
in ``Okay, I'll try this one.'' or ``How do we get back down?''.

%% file: sections/tables/leakage.tex
\begin{table}[htb]
\centering
\caption{\textbf{Source-game leakage} on all 2{,}470 test clips, without a word cap. Mentions: names of the
test game. Leaked: Baldur's Gate~3 names that exist neither in the test game nor in its wiki. Leakage: share of
leaked names among all names said.}
\label{tab:leakage}
\small
\setlength{\tabcolsep}{2pt}
\renewcommand{\arraystretch}{1.1}
\begin{tabularx}{\linewidth}{@{}l *{3}{>{\centering\arraybackslash}X}@{}}
\toprule
Method & Mentions$\uparrow$ & Leaked$\downarrow$ & Leakage$\downarrow$ \\
\midrule
VideoLLM-online~\cite{chen2024videollm} & 372  & 14  & 3.6\% \\
\dossier                                & 676  & 1   & 0.1\% \\
LiveCC-7B-Instruct~\cite{chen2025live}  & 217  & 2   & 0.9\% \\
\dossier                                & 867  & 4   & 0.5\% \\
MMDuet~\cite{wang2025videollm}          & 298  & 0   & 0.0\% \\
\dossier                                & 99   & 1   & 1.0\% \\
StreamingVLM~\cite{xu2026streamingvlm}  & 1{,}271 & 265 & 17.3\% \\
\dossier                                & 1{,}188 & 205 & 14.7\% \\
MMDuet2$^\ddagger$~\cite{wang2026mmduet2} & 147 & 4  & 2.6\% \\
\dossier                                & 135  & 2   & 1.5\% \\
Proact-VL~\cite{yan2026proact}          & 1{,}811 & 200 & 9.9\% \\
\dossier                                & 2{,}533 & 170 & 6.3\% \\
\midrule
\rowcolor{oursbg}
\textbf{StreamDuo (ours)}               & \textbf{3{,}329} & 34 & 1.0\% \\
\bottomrule
\end{tabularx}
\end{table}

%% file: sections/tables/efficiency.tex
\begin{table}[htb]
\centering
\caption{\textbf{Latency} on one NVIDIA H100 GPU, one request at a time. Research agent: seconds per answer over
200 test moments. Commentator: seconds of compute per second of video over 20 test clips, including frame
decoding.}
\label{tab:latency}
\small
\setlength{\tabcolsep}{2pt}
\renewcommand{\arraystretch}{1.1}
\begin{tabularx}{\linewidth}{@{}l *{4}{>{\centering\arraybackslash}X}@{}}
\toprule
Stage & Mean & Median & P90 & Max \\
\midrule
\multicolumn{5}{@{}l}{\textit{Research agent (s per answer)}} \\
Propose search keys      & 0.322 & 0.272 & 0.594 & 0.628 \\
BM25 wiki search         & 0.002 & 0.001 & 0.005 & 0.011 \\
Pick entries             & 0.082 & 0.091 & 0.114 & 0.188 \\
\rowcolor{oursbg}
Total                    & 0.405 & 0.363 & 0.643 & 0.713 \\
\midrule
\multicolumn{5}{@{}l}{\textit{Commentator (s per video second)}} \\
\rowcolor{oursbg}
Total                    & 0.256 & 0.242 & 0.306 & 0.332 \\
\bottomrule
\end{tabularx}
\end{table}